\documentclass[10pt,twocolumn,letterpaper]{article}

\usepackage[pagenumbers]{cvpr} % To force page numbers, e.g. for an arXiv version

\definecolor{cvprblue}{rgb}{0.21,0.49,0.74}
\usepackage[pagebackref,breaklinks,colorlinks,allcolors=cvprblue]{hyperref}

\definecolor{cvprblue}{rgb}{0.21,0.49,0.74}
\definecolor{aliceblue}{rgb}{0.87, 0.92, 0.96}
\usepackage[pagebackref,breaklinks,colorlinks,allcolors=cvprblue]{hyperref}
\usepackage{graphicx}
\usepackage{amsmath}
\usepackage{amssymb}
\usepackage{amsfonts}
\usepackage{algorithm}
\usepackage{algpseudocode}
\usepackage{algorithm}
\usepackage{algpseudocode}
\usepackage{multirow}
\usepackage{booktabs}
\usepackage{graphicx}
\usepackage[table]{xcolor}

\def\confName{CVPR}
\def\confYear{2027}

\title{Beyond Language Optimization: Enhancing Visual Representations for Radiology Report Generation}

\author{Rui Zuo\\
Zhejiang University\\
Hangzhou 310027, Zhejiang, China\\
{\tt\small ruizuo@zju.edu.cn}
\and
Qinyue Tong\\
Zhejiang University\\
Hangzhou 310027, Zhejiang, China\\
{\tt\small qinyuetong@zju.edu.cn}
\and
Ziqian Lu\\
Zhejiang Sci-Tech University\\
Hangzhou 310018, Zhejiang, China\\
{\tt\small ziqianlu@zstu.edu.cn}
\and
Zheming Lu\\
Zhejiang University\\
Hangzhou 310027, Zhejiang, China\\
{\tt\small zheminglu@zju.edu.cn}
}

\begin{document}
\maketitle
\begin{abstract}
Accurate chest X-ray radiology report generation (RRG) requires recognizing clinically relevant abnormalities and accurately describing them. Although Group Relative Policy Optimization (GRPO) improves the clinical accuracy of generated reports, our analyses suggest limited improvements in visual representations for abnormality recognition. To address this limitation, we propose a \textbf{selective regional visual distillation framework} that enhances visual representations of challenging abnormalities while preserving existing visual capabilities. Specifically, we train a region-aware teacher with targeted supervision on challenging anatomical regions to acquire complementary visual knowledge. Rather than transferring all teacher features, we verify their utility through regional feature replacement and select only those that improve the original student's performance. We further introduce a \textbf{S}patially \textbf{G}ated \textbf{R}esidual \textbf{A}dapter (\textbf{SGRA}) to distill these verified features while minimizing interference with the student's existing visual representations. Experiments on MIMIC-CXR and IU-Xray demonstrate improved clinical report accuracy and region-level abnormality recognition across different multimodal large language model backbones, while largely preserving performance on non-target findings.
\end{abstract}    
\section{Introduction}
\label{sec:intro}

Radiology report generation (RRG) aims to automatically produce textual reports describing findings in medical images, thereby assisting radiologists in report drafting and reducing clinical workload~\cite{jing2018automatic, acosta2024impact}. 
For chest X-ray reporting, earlier approaches typically employ task-specific encoder–decoder models to learn the mapping from chest X-ray images to report text~\cite{chen2020generating, chen2021cross, huang2023kiut, liu2021exploring, wang2023metransformer}, achieving substantial progress in language generation quality and clinical consistency~\cite{qin2022reinforced, liu2019clinically, tanida2023interactive}. 
However, these methods largely follow a fixed image-to-report generation paradigm and offer limited support for open-ended natural language instructions, making it difficult to flexibly guide report generation through textual prompts~\cite{ranjit2023retrieval}.

\begin{figure}[t]
    \centering
    \vspace{-16pt}
    \includegraphics[width=\columnwidth]{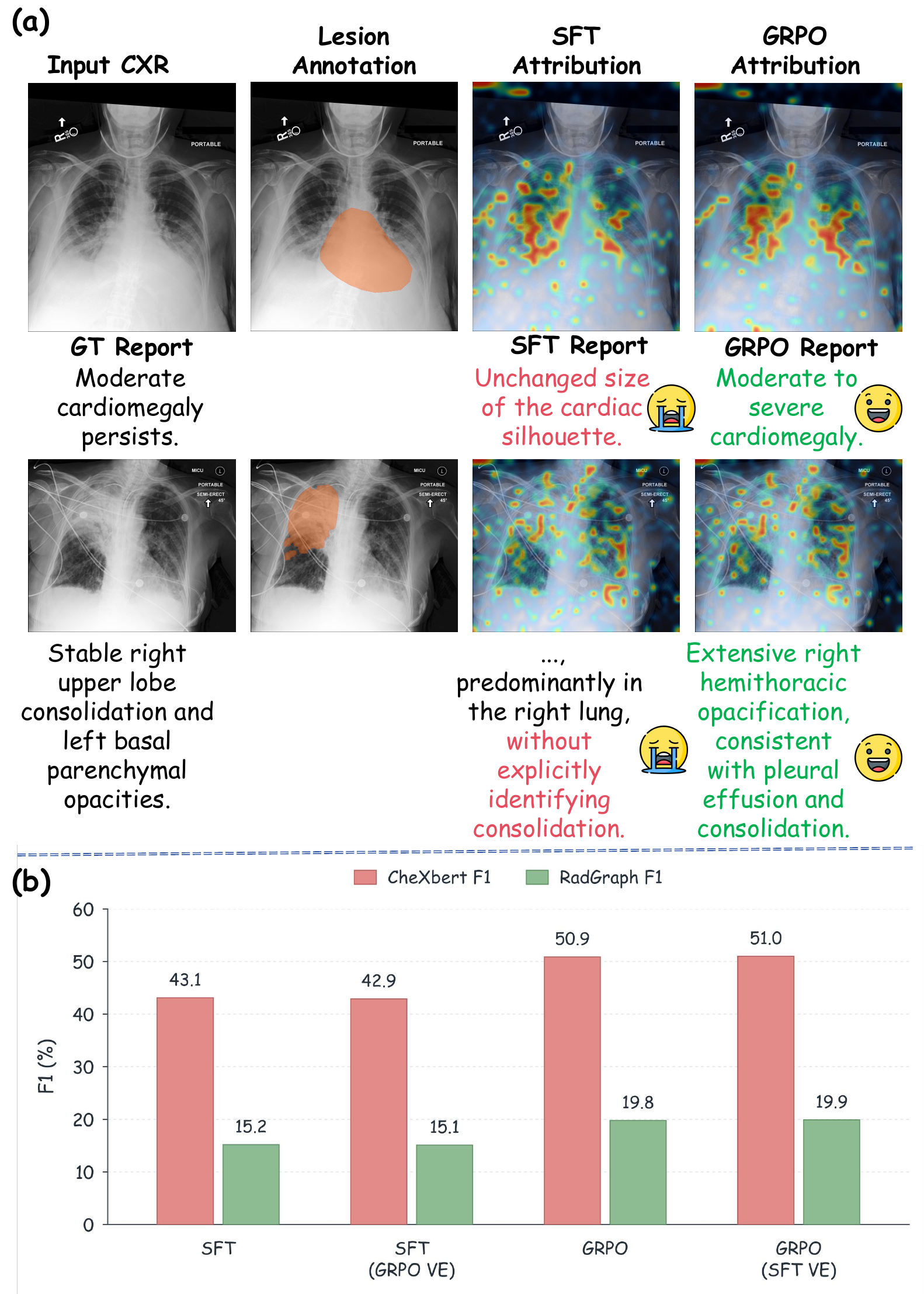}
    \caption{
        \textbf{Clinical performance and visual attribution before and after GRPO.}
        We focus on Qwen3-VL, as its larger clinical gains from GRPO compared with MedGemma make the effects of GRPO more evident. 
        \textbf{(a)} Representative CXR cases show similar disease-region attribution patterns despite corrected clinical predictions after GRPO. 
        \textbf{(b)} Exchanging visual encoders between SFT and GRPO models has little effect on CheXbert F1 and RadGraph F1. 
        The multimodal projector remains frozen in both stages.
        Parentheses indicate the source of the substituted visual encoder (VE), while the leading model name indicates the source of the remaining components. 
        GT denotes ground truth.
    }
    \label{fig:intro_vis}
    \vspace{-5pt}
\end{figure}

In contrast, multimodal large language models (MLLMs) enable flexible instruction-guided report generation and are commonly adapted to RRG through supervised fine-tuning (SFT)~\cite{liu2023visual,hyland2023maira,zambrano2025clinically}.
However, SFT optimizes the likelihood of reference reports rather than directly optimizing the clinical correctness of generated reports~\cite{gundersen2026radvlm}. 
This objective mismatch motivates reward-based optimization, which enables clinical quality measures to guide model updates~\cite{wang2025mrg}. 
In particular, emerging RRG approaches adopt Group Relative Policy Optimization (GRPO)~\cite{zhang2025editgrpo}, which uses relative rewards within groups of sampled reports and avoids the separate value model required by PPO-based reinforcement learning~\cite{shao2024deepseekmath}.
Despite its success in verifiable tasks such as mathematical reasoning and code generation~\cite{guo2025deepseek}, GRPO faces additional challenges in RRG, as a report contains multiple clinical findings while a single report-level reward can obscure which statements are correct or erroneous~\cite{chen2026orapo}.
To mitigate this issue, existing approaches incorporate clinical fact-based reward formulations~\cite{chen2026orapo} and sentence-level corrective feedback~\cite{zhang2025editgrpo} to improve the clinical accuracy of generated reports.
However, accurate RRG relies on two complementary capabilities: extracting clinically relevant visual information from radiographs and faithfully describing the corresponding clinical findings.
Despite the improvements achieved by these approaches, it remains unclear whether their gains stem from improvements in the visual features extracted from radiographs or more effective utilization of existing visual information for report generation.
This raises a key question: \textbf{does better reporting reflect improved abnormality detection or better descriptions of already recognized findings?}

To investigate this question, we conduct controlled experiments with Qwen3-VL~\cite{bai2025qwen3} on MIMIC-CXR~\cite{johnson2019mimic}. 
Our experiments show GRPO improves the clinical accuracy of reports generated by the SFT-initialized model, while unlocking and freezing the visual encoder yield nearly identical clinical performance. 
To investigate whether these improvements reflect better visual feature extraction, we sample 100 test cases comprising 178 disease instances, of which 44 change from incorrect to correct predictions after GRPO, while 20 change in the opposite direction.
Despite corrected clinical predictions after GRPO, report-level gradient-based attribution maps show highly similar local patterns within the corresponding disease regions for over 70\% of the 44 corrected instances. 
Examples are shown in \cref{fig:intro_vis}(a), prompting us to further investigate through visual encoder exchange whether improved report quality is accompanied by stronger visual representations.
As shown in \cref{fig:intro_vis}(b), exchanging the visual encoders between the SFT and GRPO models leaves both models' CheXbert F1~\cite{smit2020combining} and RadGraph F1~\cite{jain2021radgraph} nearly unchanged, with neither model benefiting
appreciably from the GRPO visual encoder.
Together with the attribution results, this suggests that GRPO offers limited improvement in visual representations relevant to abnormality recognition, despite improving clinical report quality.

\begin{figure}[t]
    \centering
    \vspace{-16pt}
    \includegraphics[width=\columnwidth]{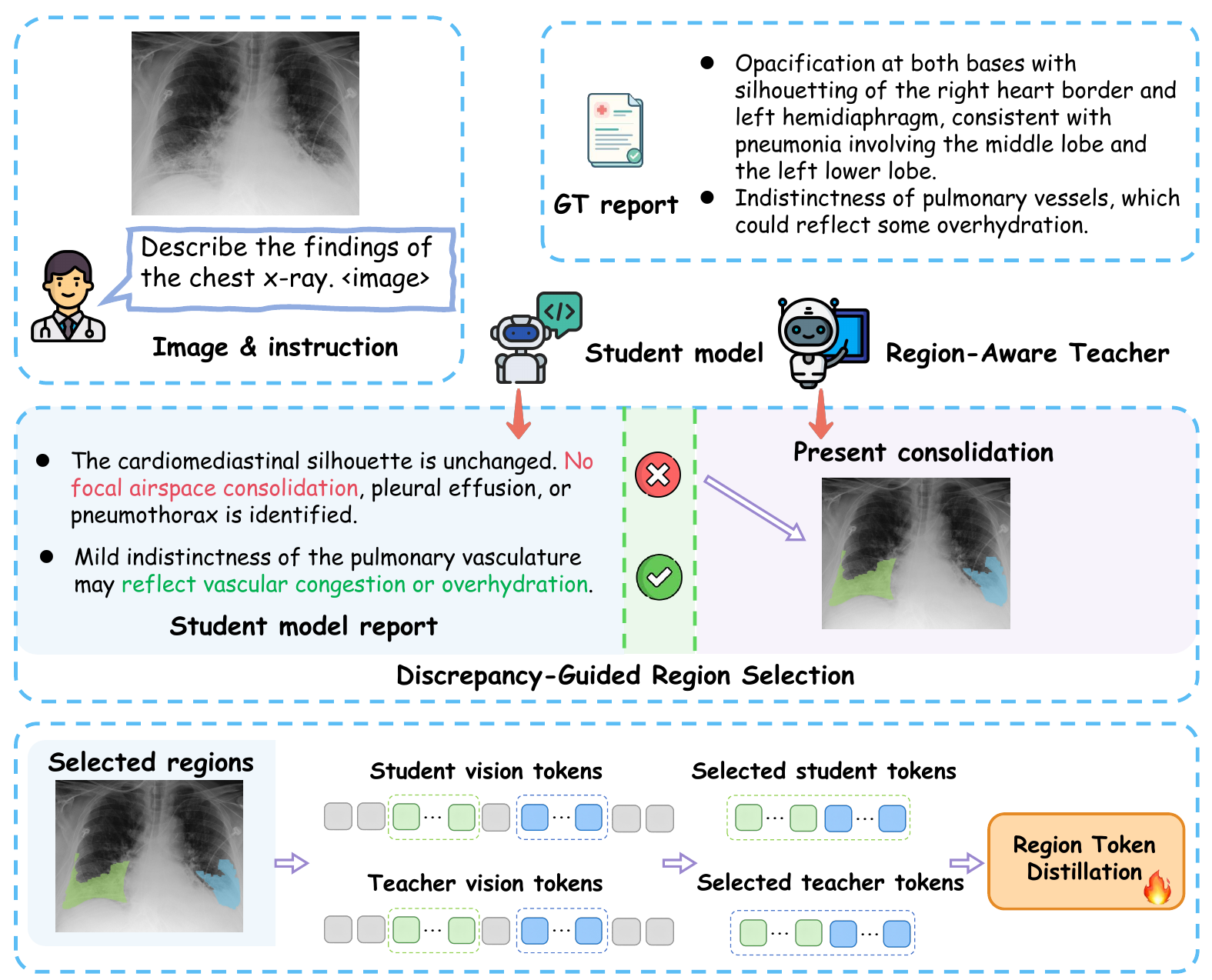}
    \caption{
        The region-aware teacher, trained with region-level supervision, provides targeted guidance for enhancing the student's visual representations in challenging regions.
        When the student incorrectly predicts a disease while the teacher correctly identifies it in the corresponding image region, that region is selected for distillation after passing regional feature utility verification. 
        The visual tokens corresponding to the selected region are then extracted from both models, and the student's tokens are aligned with their teacher counterparts to improve the visual representations associated with the student's prediction errors. 
    }
    \label{fig:region_distillation}
    \vspace{-5pt}
\end{figure}

% We hypothesize that the limited visual improvements observed in CXR report generation stem from the lack of direct guidance for visual representation learning in text-based reward optimization. 
We hypothesize that the limited visual improvements in CXR report generation stem from the lack of direct guidance for visual representation learning in text-based reward optimization.
% Although rewards can distinguish correct from incorrect descriptions of specific abnormalities, they do not explicitly constrain the visual representations of the corresponding lesion regions, potentially favoring language-side optimization that makes better use of existing visual information. 
Although rewards can distinguish correct from incorrect descriptions of specific abnormalities, they do not explicitly constrain visual representations of the corresponding lesion regions, potentially favoring language-side optimization that better leverages existing visual information. 
To address this limitation, we seek to directly enhance visual representations associated with abnormality recognition.
Given the relatively consistent anatomical structure of chest X-rays, region-level supervision provides a natural way to guide visual representation learning in disease-associated regions. 
Rather than optimizing visual representations indiscriminately, we focus on challenging regions where the model predicts incorrectly or struggles to distinguish confusable and coexisting abnormalities. 
However, optimizing visual representations on a limited set of challenging cases may introduce prediction bias and perturb representations in other regions, potentially improving recognition of difficult abnormalities at the expense of performance on normal cases and in unaffected regions.

To address this challenge, we decouple the enhancement of regional visual representations from direct updates to the original model. 
Specifically, we train a region-aware teacher using targeted supervision on annotated regions in challenging cases, allowing it to acquire complementary visual knowledge even without achieving superior overall performance. 
As illustrated in \cref{fig:region_distillation}, rather than transferring all teacher representations, we identify cases where the teacher predicts correctly but the original model does not, further verify whether the corresponding regional features improve the original model's performance, and selectively distill the validated visual tokens into the student model.
This selective distillation aims to improve the recognition of challenging abnormalities in targeted regions while preserving the model's existing full-image visual representations and performance on normal cases. 
Furthermore, to limit changes to visual representations outside the targeted regions, we introduce a Spatially Gated Residual Adapter (\textbf{SGRA}) that learns spatially adaptive corrections to the original visual features. 
Extensive experiments demonstrate that our method enhances visual representations in regions corresponding to challenging abnormalities, improves the clinical accuracy of generated reports, and largely preserves the model's existing visual capabilities. 
Our main contributions are summarized as follows:
\begin{itemize}
    \item We propose a selective regional visual
    distillation framework that validates teacher
    features through regional
    replacement before distillation, enabling
    student-beneficial visual knowledge transfer
    for challenging abnormalities.

    \item We design a Spatially Gated Residual
    Adapter (SGRA) with complementary feature
    distillation and preservation objectives to
    improve previously incorrect abnormality
    recognition while minimizing interference
    with the student's existing visual capabilities.

    \item Extensive experiments demonstrate the
    effectiveness of our approach across different
    MLLM backbones for RRG, achieving state-of-the-art performance on selected evaluation metrics across multiple benchmarks.
\end{itemize}

\section{Related Works}
\label{sec:relatedworks}

%-------------------------------------------------------------------------
\subsection{Radiology Report Generation}
Early radiology report generation (RRG) methods adopt task-specific encoder–decoder architectures to transform visual features extracted from medical images into textual reports~\cite{chen2020generating, chen2021cross}. 
Subsequent studies incorporate medical knowledge, cross-modal alignment, and anatomical information to strengthen the correspondence between visual findings and clinical descriptions, improving report accuracy~\cite{liu2021exploring, huang2023kiut, wang2023metransformer, jin2024promptmrg, bu2024instance, park2025dart, wang2025cxpmrg}. 
More recently, multimodal large language models (MLLMs) enable flexible instruction-guided report generation through supervised fine-tuning~\cite{hyland2023maira, zambrano2025clinically, wang2023r2gengpt, liu2024bootstrapping}, with some approaches leveraging region-level supervision and anatomical cues to improve abnormality recognition and visual grounding~\cite{tanida2023interactive, Messina_2026_CVPR, bannur2024maira, gu2025orid, chen2024act}. 
Despite these advances, existing methods primarily focus on improving report generation quality, with limited exploration of how to selectively enhance visual representations for abnormalities that remain difficult to recognize. 
In contrast, our approach targets these challenging abnormalities through selective regional visual distillation while preserving the model's existing visual capabilities.

\subsection{Group Relative Policy Optimization}
Group Relative Policy Optimization (GRPO) efficiently optimizes policies using relative rewards within groups of sampled responses without requiring a separate value model~\cite{shao2024deepseekmath}. 
Following its success in complex reasoning tasks~\cite{guo2025deepseek, yu2026dapo}, GRPO has been extended to multimodal reasoning and visual perception~\cite{zhang2025r1,liu2025visual}, as well as radiology report generation (RRG) to improve the clinical accuracy of generated reports~\cite{wang2025mrg,gundersen2026radvlm,liu2026scaling}.
However, the presence of multiple clinical findings in a single report makes report-level rewards insufficient for providing precise feedback on individual abnormalities. 
Existing approaches introduce clinical fact-based rewards and fine-grained corrective feedback to better guide the optimization of report generation~\cite{zhang2025editgrpo, chen2026orapo, akhter2026clinically, chen2026precision, zhao2026aligning}. 
Nevertheless, these reward-based approaches do not directly constrain reliable visual representations of the corresponding abnormalities. 
Our method complements GRPO by explicitly enhancing visual representations of challenging abnormalities through more utility-verified regional knowledge distillation.

\section{Methodology}

\begin{figure*}[t]
    \centering
    \vspace{-16pt}
    \includegraphics[width=\textwidth]{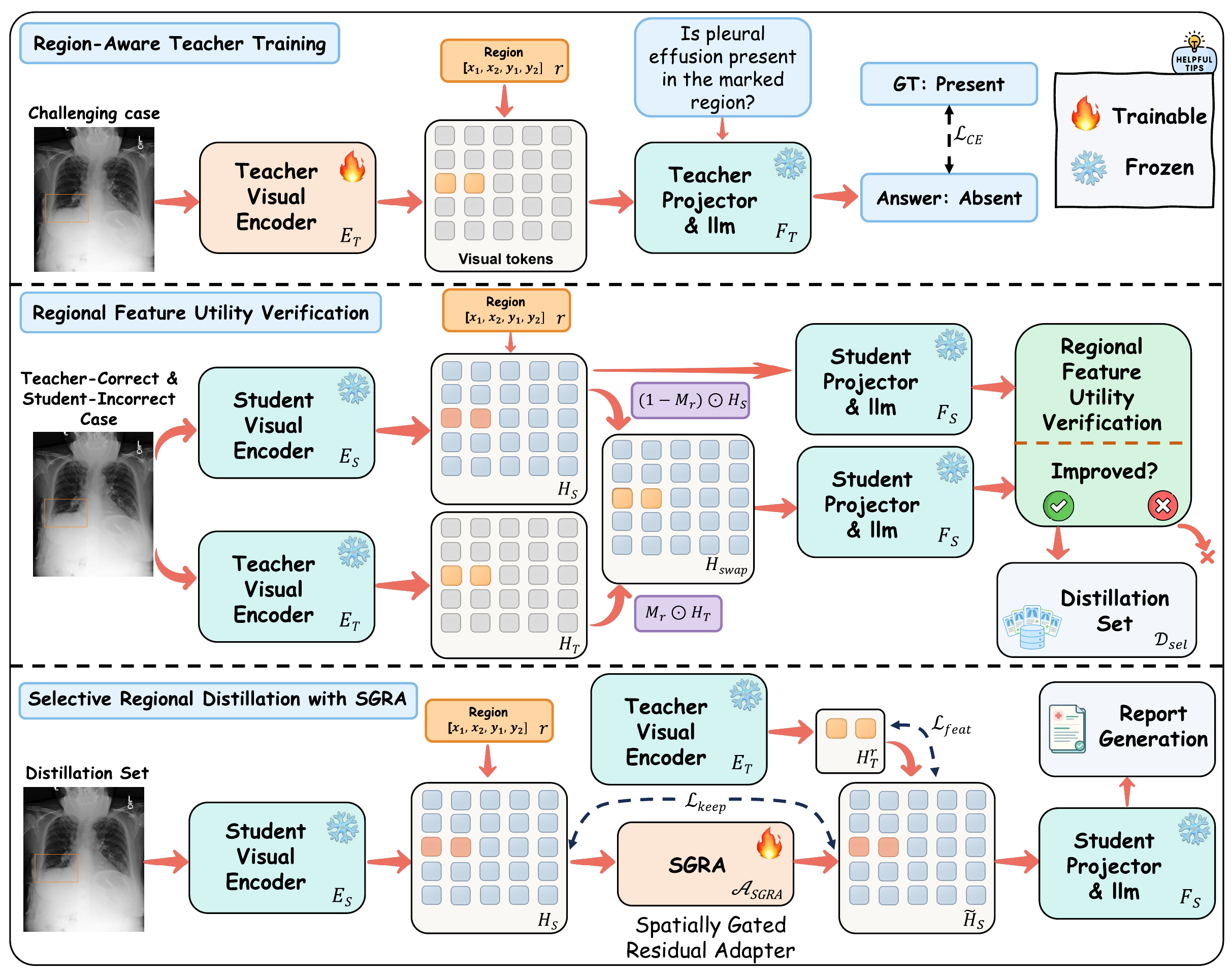}
    \caption{Overview of the proposed selective regional visual distillation framework. 
    \textbf{Region-Aware Teacher Training:} 
    A region-aware teacher is trained with region-specific questions and anatomical annotations to enhance visual representations of disease-associated regions in challenging cases. \textbf{Regional Feature Utility Verification:} 
    Teacher features are substituted into the corresponding student regions, and the resulting changes in the student's correct-answer scores are used to identify beneficial regional features for distillation. 
    \textbf{Selective Regional Distillation with SGRA:} 
    A Spatially Gated Residual Adapter (SGRA) learns from the selected teacher features through regional feature distillation, while a feature preservation loss constrains changes to non-target regions.
    }
    \label{fig:method}
    \vspace{-18pt}
\end{figure*}

To address the limited improvement in visual representations following SFT and GRPO in the RRG task, we propose a selective regional visual distillation framework that explicitly enhances visual representations of previously misidentified abnormalities while preserving its existing visual capabilities. 
As illustrated in \cref{fig:method}, our framework consists of three stages: region-aware teacher training, regional feature utility verification, and selective regional distillation with SGRA.

Starting from an SFT- and GRPO-trained model, we initialize a teacher and a student with identical weights.
First, we train the teacher with targeted supervision on disease-specific annotated regions in challenging cases, enabling it to learn more discriminative visual representations of the corresponding abnormalities without requiring an overall improvement in its visual capabilities (Sec.~\ref{sec:method-1}).
Next, we identify cases where the teacher predicts correctly but the student does not and further verify whether the corresponding teacher features improve the student's performance through regional feature replacement (Sec.~\ref{sec:method-2}).
Finally, we train SGRA to distill these verified regional features into the student while minimizing the impact on its existing visual representations (Sec.~\ref{sec:method-3}).

\subsection{Region-aware teacher training}
\label{sec:method-1}

We construct a region-level question-answering dataset from a curated single-image subset of MIMIC-CXR~\cite{johnson2019mimic}, leveraging anatomical bounding boxes and report-derived finding annotations from Chest ImaGenome~\cite{wu2021chest}, with patient-disjoint training, validation, and test splits. 
Each example pairs a radiograph and an anatomical region with a disease-specific presence question targeting one of fo abnormalities: atelectasis, consolidation, cardiomegaly, and pleural effusion.

We denote the resulting teacher training dataset as
\begin{equation}
    \mathcal{D}_T = \{(I_i, r_i, c_i, q_i, y_i)\}_{i=1}^{N},
    \label{eq:teacher_dataset}
\end{equation}
where $I_i$, $r_i$, $c_i$, $q_i$, and $y_i$ represent the radiograph, anatomical region, target abnormality, region-specific question, and binary presence label, respectively.
Binary labels are assigned based on explicit positive or negative evidence, including applicable regional normality statements, while uncertain, conflicting, and unmentioned findings are excluded. 
We identify challenging examples based on either incorrect regional predictions by the fixed student model under the same question-answering setup or potentially confounding findings supported by the report, such as the presence of another abnormality in the target region despite the target disease being absent, or the coexistence of multiple positive findings within the same region.

We use these challenging examples to train a region-aware teacher that can complement the original student's visual representations in disease-associated anatomical regions. Both models are initialized from the same SFT- and GRPO-trained model:
\begin{equation}
    \mathcal{M}_T^{(0)} = \mathcal{M}_S = \mathcal{M}_0
    = (\mathcal{E}_0, \mathcal{F}_0),
    \label{eq:model_init}
\end{equation}
where $\mathcal{E}_0$ denotes the visual encoder and $\mathcal{F}_0$ represents the combined multimodal projector and LLM.
As illustrated in the \textit{Region-Aware Teacher Training} stage of \cref{fig:method}, we optimize only the teacher visual encoder \(\mathcal{E}_T\) while keeping the multimodal projector and LLM frozen.

Specifically, we map each annotated anatomical region $r_i$ onto the spatial grid of visual tokens, obtaining a binary region mask $M_{r_i}$:
\begin{equation}
    M_{r_i}[j] =
    \begin{cases}
        1, & \text{if } A_j \cap r_i \neq \varnothing,\\
        0, & \text{otherwise},
    \end{cases}
    \label{eq:region_mask}
\end{equation}
where $A_j$ denotes the image area corresponding to the
$j$-th visual token. 
We inject a region-marker embedding into the visual tokens selected by $M_{r_i}$ and insert a special token \texttt{<region\_target>} into the textual prompt, whose input embedding shares the same marker component.
Notably, the region-marker embeddings are pretrained separately to enable the model to recognize and associate the region markers across visual and textual modalities, and are subsequently shared by the teacher and student and kept frozen during teacher training. 
Then, we train the teacher to predict the binary answer $y_i \in \{\mathrm{Present}, \mathrm{Absent}\}$ for the corresponding question $q_i$ by minimizing the negative log-likelihood of the ground-truth answer:
\begin{equation}
    \mathcal{L}_T =
    -\mathbb{E}_{(I,r,c,q,y)\sim\mathcal{D}_T}
    \left[
        \log P_{\mathcal{M}_T}(y \mid I, M_r, q)
    \right].
    \label{eq:teacher_loss}
\end{equation}
Although the teacher is trained to better recognize abnormalities in the corresponding regions of challenging cases, whether its learned regional visual features can benefit the original student remains uncertain. We therefore assess their utility in the following stage.

\subsection{Regional feature utility verification}
\label{sec:method-2}

To assess whether the region-aware teacher can provide useful visual representations for the student in disease-associated regions of challenging cases, we evaluate its regional features through targeted feature replacement.
As illustrated in the \textit{Regional Feature Utility Verification} stage of \cref{fig:method}, starting from cases where the teacher predicts correctly but the student does not, we extract their respective visual representations from the same radiograph $I$:
\begin{equation}
    H_T = \mathcal{E}_T(I), \quad
    H_S = \mathcal{E}_S(I),
    \label{eq:visual_features}
\end{equation}
We then replace the student's visual tokens in the target region
with the corresponding teacher tokens, while preserving the
student's representations in all other regions:
\begin{equation}
    H_{\mathrm{swap}} =
    (1-M_r)\odot H_S + M_r\odot H_T,
    \label{eq:feature_swap}
\end{equation}
where $M_r$ is the binary region mask in \cref{eq:region_mask}, and $\odot$ denotes element-wise multiplication.
The resulting features $H_{\mathrm{swap}}$ and the student features $H_S$ are evaluated using the same frozen student projector and LLM under an identical region-specific question $q$. 
The same frozen region markers are applied during QA scoring, while visual features used for replacement and distillation remain unmarked. 
We measure the student's support for the correct answer as:
\begin{equation}
    \mathcal{S}(H) =
    \log P_{\mathcal{F}_S}(y \mid H, q)
    - \log P_{\mathcal{F}_S}(\bar{y} \mid H, q),
    \label{eq:answer_score}
\end{equation}
where $y$ and $\bar{y}$ denote the ground-truth answer and its opposite, respectively, and $\mathcal{F}_S$ represents the frozen student projector and LLM.
The utility gain of regional feature replacement is calculated as:
\begin{equation}
    u = \mathcal{S}(H_{\mathrm{swap}})
        - \mathcal{S}(H_S),
    \label{eq:utility_gain}
\end{equation}
We retain cases with $u > \tau$, where $\tau$ is a predefined utility threshold, to construct the selected distillation set $\mathcal{D}_{\mathrm{sel}}$.

\subsection{Selective regional distillation with SGRA}
\label{sec:method-3}

Using the selected distillation set $\mathcal{D}_{\mathrm{sel}}$,
we aim to transfer useful regional visual knowledge from the teacher to the student while minimizing changes to the student's representations in non-target regions.
As shown in the \textit{Selective Regional Distillation
with SGRA} stage of \cref{fig:method}, we freeze the student
visual encoder, projector, and LLM, and introduce a trainable
\textbf{Spatially Gated Residual Adapter (SGRA)} between the visual
encoder and projector to learn from the teacher's regional
visual features. 
Given the student visual representations $H_S=\{h_i\}_{i=1}^{N_v}$, SGRA produces the adapted representations $\widetilde{H}_S=\{\widetilde{h}_i\}_{i=1}^{N_v}$ through a spatially gated residual transformation:
\begin{equation}
    \widetilde{h}_i =
    h_i + g_i(H_S)\cdot\Delta h_i(H_S),
    \label{eq:sgra}
\end{equation}
where $g_i(H_S)\in[0,1]$ controls the adaptation strength at the $i$-th visual token, and $\Delta h_i(H_S)$ denotes the learned residual feature.
Specifically, SGRA projects the visual features into a low-dimensional space and combines local spatial features extracted by depthwise convolution with global context obtained through average pooling. 
The fused features are then used to predict token-wise gating weights and residual corrections.

To transfer the verified regional knowledge, we introduce a feature distillation loss $\mathcal{L}_{\mathrm{feat}}$ that aligns the adapted student features with the teacher features within the selected target regions.
Meanwhile, a feature preservation loss $\mathcal{L}_{\mathrm{keep}}$ constrains changes to the student's representations outside these regions.
The losses and overall objective are defined as:
\begin{equation}
\label{eq:regional_distillation_loss}
\begin{split}
\mathcal{L}_{\mathrm{feat}}
&= \mathbb{E}_{\mathcal{D}_{\mathrm{sel}}}
\frac{\sum_i M_r[i]\|\widetilde{h}_i-h_{T,i}\|_2^2}
{C\sum_i M_r[i]},\\
\mathcal{L}_{\mathrm{keep}}
&= \mathbb{E}_{\mathcal{D}_{\mathrm{sel}}}
\frac{\sum_i (1-M_r[i])\|\widetilde{h}_i-h_i\|_2^2}
{C\sum_i (1-M_r[i])},\\
\mathcal{L}
&= \lambda_{\mathrm{feat}}\mathcal{L}_{\mathrm{feat}}
+\lambda_{\mathrm{keep}}\mathcal{L}_{\mathrm{keep}}.
\end{split}
\end{equation}
where $h_{T,i}$ denotes the teacher feature at the $i$-th visual token, $C$ is the feature dimension, and $\lambda_{\mathrm{feat}}$ and $\lambda_{\mathrm{keep}}$ balance the two losses.

Notably, although SGRA is trained with region-level supervision, it operates solely on the student's visual features during inference, without requiring the teacher, anatomical region annotations, or region-specific questions.
By selectively transferring useful regional features from the teacher, our method aims to improve the recognition of previously misidentified abnormalities while minimizing interference with the student's previously learned visual capabilities.

\section{Experiments}

\begin{table*}[t]
\centering
\caption{Comparison with state-of-the-art methods on MIMIC-CXR and IU-Xray. All methods are trained on MIMIC-CXR for RRG and directly evaluated on IU-Xray without any further fine-tuning. Our method uses only MIMIC-CXR for task-specific training, while some baselines may include IU-Xray in their pretraining. Micro-F1 and Macro-F1 are computed using CheXbert, while RG-F1 denotes RadGraph Partial F1. BLEU-1/2/3/4 and METEOR are computed at the corpus level, whereas ROUGE-L and RadGraph Partial F1 are averaged over individual reports. Best and second-best results are highlighted in \textbf{bold} and \underline{underlined}, respectively.}
\label{tab:sota}
\setlength{\tabcolsep}{4pt}
\renewcommand{\arraystretch}{1.08}
\resizebox{\textwidth}{!}{
\begin{tabular}{cccccccccccc}
\toprule
\multirow{2}{*}{Dataset} &
\multirow{2}{*}{Method} &
\multirow{2}{*}{Venue} &
\multicolumn{6}{c}{NLG Metrics $\uparrow$} &
\multicolumn{3}{c}{Clinical Metrics $\uparrow$} \\
\cmidrule(lr){4-9}
\cmidrule(lr){10-12}
& & &
B-1 & B-2 & B-3 & B-4 &
METEOR & ROUGE-L & Micro-F1 & Macro-F1 & RG-F1 \\
\midrule

\multicolumn{1}{c|}{\multirow{6}{*}{MIMIC-CXR}}
& RGRG~\cite{tanida2023interactive} & CVPR'23
& 34.36 & 20.83 & \underline{13.36} & \underline{8.73} & 16.26 & 25.04
& 54.25 & 29.77 & 22.71 \\
\multicolumn{1}{c|}{}
& CURV~\cite{wang2026curv} & NeurIPS'25
& 30.06 & 15.86 & 10.68 & 6.38 & 12.02 & 23.16
& 41.60 & 18.57 & \underline{22.86} \\
\multicolumn{1}{c|}{}
& MambaXray-VL~\cite{wang2025cxpmrg} & CVPR'25
& 36.36 & 20.77 & 12.61 & 7.96 & 14.88 & 24.85
& 48.97 & 22.40 & 19.61 \\
\multicolumn{1}{c|}{}
& REVA-PO~\cite{guo2026revapo} & ECCV'26
& 32.74 & 20.23 & 12.96 & 8.42 & \underline{16.60} & \underline{26.35}
& 57.39 & 30.86 & 21.24 \\
\multicolumn{1}{c|}{}
& FairLLaVA~\cite{bhosale2026fairllava} & CVPR'26
& \underline{36.71} & \underline{21.03} & 13.15 & 8.67 & 13.93 & 25.93
& \underline{59.12} & \underline{34.64} & 21.10 \\
\multicolumn{1}{c|}{}
& \cellcolor{aliceblue!60}\textbf{Ours}
& \cellcolor{aliceblue!60}--
& \cellcolor{aliceblue!60}\textbf{37.89}
& \cellcolor{aliceblue!60}\textbf{24.25}
& \cellcolor{aliceblue!60}\textbf{16.52}
& \cellcolor{aliceblue!60}\textbf{11.45}
& \cellcolor{aliceblue!60}\textbf{18.48}
& \cellcolor{aliceblue!60}\textbf{30.98}
& \cellcolor{aliceblue!60}\textbf{63.02}
& \cellcolor{aliceblue!60}\textbf{36.27}
& \cellcolor{aliceblue!60}\textbf{28.50} \\
\midrule

\multicolumn{1}{c|}{\multirow{6}{*}{IU-Xray}}
& RGRG~\cite{tanida2023interactive} & CVPR'23
& 36.49 & 23.97 & 16.60 & 11.66 & 20.85 & 31.35
& 60.24 & \underline{18.24} & 30.01 \\
\multicolumn{1}{c|}{}
& CURV~\cite{wang2026curv} & NeurIPS'25
& 22.37 & 14.26 & 9.17 & 6.20 & 20.03 & 25.34
& 36.10 & 12.86 & 31.57 \\
\multicolumn{1}{c|}{}
& MambaXray-VL~\cite{wang2025cxpmrg} & CVPR'25
& 30.02 & 17.77 & 11.01 & 7.14 & 18.53 & 27.23
& \underline{61.37} & 17.57 & 23.41 \\
\multicolumn{1}{c|}{}
& REVA-PO~\cite{guo2026revapo} & ECCV'26
& 29.15 & 18.47 & 12.06 & 8.17 & \textbf{22.09} & 28.69
& 43.41 & 14.96 & 30.68 \\
\multicolumn{1}{c|}{}
& FairLLaVA~\cite{bhosale2026fairllava} & CVPR'26
& \underline{46.42} & \underline{29.15} & \underline{19.56} & \underline{13.50} & 21.39 & \textbf{36.91}
& 55.49 & 17.85 & \underline{35.91} \\
\multicolumn{1}{c|}{}
& \cellcolor{aliceblue!60}\textbf{Ours}
& \cellcolor{aliceblue!60}--
& \cellcolor{aliceblue!60}\textbf{51.94}
& \cellcolor{aliceblue!60}\textbf{32.23}
& \cellcolor{aliceblue!60}\textbf{21.21}
& \cellcolor{aliceblue!60}\textbf{14.07}
& \cellcolor{aliceblue!60}\underline{21.74}
& \cellcolor{aliceblue!60}\underline{34.25}
& \cellcolor{aliceblue!60}\textbf{74.49}
& \cellcolor{aliceblue!60}\textbf{25.35}
& \cellcolor{aliceblue!60}\textbf{39.14} \\
\bottomrule
\end{tabular}
}
\end{table*}

\subsection{Experimental Setup}
\label{sec:experiments-1}
\noindent\textbf{Datasets.}
We evaluate our method on two widely used RRG benchmarks, MIMIC-CXR~\cite{johnson2019mimic} and IU-Xray~\cite{demner2016preparing}, with all training conducted exclusively on MIMIC-CXR. 
We construct patient-disjoint training, validation, and test splits from MIMIC-CXR, retaining a single radiograph per sample, resulting in 95,343, 493, and 3,487 image-report pairs, respectively.
For cross-dataset evaluation, we use 590 image-report pairs from IU-Xray without further training. 

For region-level supervision, we construct a regional question-answering dataset using anatomical bounding boxes and report-derived annotations from Chest ImaGenome~\cite{wu2021chest}, focusing on four thoracic abnormalities: atelectasis, consolidation, cardiomegaly, and pleural effusion. We use 13424 chest X-ray images comprising 38552 region-level question-answer pairs to train the region-aware teacher, and 4168 images comprising 7243 region-level question-answer pairs for selective regional distillation. All training data are drawn exclusively from the MIMIC-CXR training split, with no overlap with the evaluation sets.
\par\vspace{0.3em}
\noindent\textbf{Implementation Details.}
We adopt MedGemma 1 4B~\cite{sellergren2025medgemma} as our primary backbone and evaluate the generalizability of our method on Qwen3-VL-8B-Instruct~\cite{bai2025qwen3}. 
Both the teacher and student are initialized from the SFT- and GRPO-trained model. 
Before teacher training, we pretrain the region-marker embeddings for 60 steps using a cross-entropy objective with a learning rate of $3\times10^{-2}$, while keeping the backbone frozen. 
During region-aware teacher training, only its visual encoder is updated at a learning rate of $1 \times 10^{-5}$, while the multimodal projector and language model remain frozen. 
The teacher is trained for 150 steps with a per-GPU batch size of 10. 
The utility threshold $\tau$ for selecting regional distillation samples is set to 0.05. 
During selective regional distillation, the student backbone remains frozen, and only SGRA is optimized for 150 steps with a learning rate of $1 \times 10^{-4}$ and a per-GPU batch size of 10. 
The loss weights $\lambda_{\mathrm{feat}}$ and $\lambda_{\mathrm{keep}}$ are set to 1 and 0.5, respectively. 
All models are trained on four NVIDIA RTX A6000 GPUs (48 GB each), with a training time of approximately 2.5 days, including initial model training, region-aware teacher training, and selective regional distillation. 
\par\vspace{0.3em}
\noindent\textbf{Evaluation Metrics.}
We evaluate report generation performance using both natural language generation (NLG) and clinical efficacy metrics. For NLG evaluation, we adopt BLEU-1 (B-1), BLEU-2 (B-2), BLEU-3 (B-3), BLEU-4 (B-4)~\cite{papineni2002bleu}, METEOR~\cite{banerjee2005meteor}, and ROUGE-L (R-L)~\cite{lin2004rouge}. For clinical evaluation, we report CheXbert~\cite{smit2020combining} micro-F1 and macro-F1 to assess the correctness of clinical findings, along with RadGraph~\cite{jain2021radgraph} F1 to measure the consistency of clinical entities and relations. Additionally, we evaluate region-level abnormality recognition using precision, recall, and F1 scores.

\subsection{Comparison with State-of-the-Art Methods}
\label{sec:experiments-2}

We compare our method with five representative RRG approaches, including RGRG~\cite{tanida2023interactive}, CURV~\cite{wang2026curv}, MambaXray-VL~\cite{wang2025cxpmrg}, REVA-PO~\cite{guo2026revapo}, and FairLLaVA~\cite{bhosale2026fairllava}. All methods are evaluated using MIMIC-CXR-trained checkpoints, with IU-Xray used exclusively for cross-dataset evaluation without further fine-tuning.

As shown in Table~\ref{tab:sota}, our method achieves the best results on most metrics on both MIMIC-CXR and IU-Xray, particularly in clinical accuracy as measured by CheXbert and RadGraph F1, while maintaining strong NLG performance. The consistent improvements on IU-Xray without additional fine-tuning further demonstrate the effectiveness and cross-dataset generalizability of our method.

\subsection{Ablation Studies}
\label{sec:experiments-3}
As shown in Table~\ref{tab:ablation}, vanilla regional distillation degrades performance compared with the SFT- and GRPO-trained baseline, indicating that indiscriminate transfer of teacher features can disrupt existing representations. Introducing utility-based selection substantially mitigates this degradation, while SGRA further improves clinical performance by enabling spatially adaptive feature corrections. Finally, incorporating the preservation loss yields the best overall results, highlighting the importance of retaining existing visual information during regional enhancement.

\begin{table}[t]
\centering
\caption{Ablation study on MIMIC-CXR using MedGemma 1 4B. SFT + GRPO denotes the original model without distillation. Vanilla Distillation uses a standard residual adapter to distill regional features from samples where the student is incorrect but the teacher is correct. Utility Verification follows the same setting but further filters samples based on their verified utility.}
\label{tab:ablation}
\setlength{\tabcolsep}{3.5pt}
\renewcommand{\arraystretch}{1.1}
\resizebox{\columnwidth}{!}{
\begin{tabular}{ccccccc}
\toprule
Method & Selection & SGRA & $\mathcal{L}_{\mathrm{keep}}$ &
Micro-F1 & Macro-F1 & RG-F1 \\
\midrule
SFT + GRPO          & -- & -- & -- & 60.44 & 33.27 & 27.32 \\
+ Vanilla Distillation & -- & -- & -- & 56.48 & 31.96 & 24.89 \\
+ Utility Verification & \checkmark & -- & -- & 58.47 & 34.81 & 27.03 \\
+ SGRA              & \checkmark & \checkmark & -- & 61.14 & 35.42 & 27.62 \\
\rowcolor{aliceblue!60}
\textbf{Ours}        & \checkmark & \checkmark & \checkmark
                     & \textbf{63.02} & \textbf{36.27} & \textbf{28.50} \\
\bottomrule
\end{tabular}
}
\end{table}

\subsection{Cross-backbone Generalization}
\label{sec:experiments-4}

To evaluate the generalizability of our method across different MLLM architectures, we further conduct experiments on Qwen3-VL-8B-Instruct of a different architecture. As shown in Table~\ref{tab:backbone}, our method consistently improves clinical performance on both MedGemma and Qwen3-VL across all three metrics. These results suggest that the proposed selective regional distillation is not limited to a specific backbone and can effectively improve clinical accuracy in the RRG task across different MLLM architectures.

\begin{table}[t]
\centering
\caption{Generalization of our method across different MLLM backbones on MIMIC-CXR. Qwen3-VL-8B-Instruct follows the same training and distillation pipeline described in our method.}
\label{tab:backbone}
\setlength{\tabcolsep}{4pt}
\renewcommand{\arraystretch}{1.1}
\resizebox{\columnwidth}{!}{
\begin{tabular}{ccccc}
\toprule
Backbone & Method & Micro-F1 & Macro-F1 & RG-F1 \\
\midrule
\multicolumn{1}{l|}{\multirow{2}{*}{MedGemma 1 4B}}
& SFT + GRPO & 60.44 & 33.27 & 27.32 \\
\multicolumn{1}{l|}{}
& \cellcolor{aliceblue!60}\textbf{+ Ours}
& \cellcolor{aliceblue!60}63.02
& \cellcolor{aliceblue!60}36.27
& \cellcolor{aliceblue!60}28.50 \\
\midrule
\multicolumn{1}{l|}{\multirow{2}{*}{Qwen3-VL-8B-Instruct}}
& SFT + GRPO & 50.90 & 26.78 & 19.80 \\
\multicolumn{1}{l|}{}
& \cellcolor{aliceblue!60}\textbf{+ Ours}
& \cellcolor{aliceblue!60}54.68
& \cellcolor{aliceblue!60}30.03
& \cellcolor{aliceblue!60}21.82 \\
\bottomrule
\end{tabular}
}
\end{table}

\subsection{Analysis of Regional Visual Distillation}
\label{sec:experiments-5}

We assess region-level abnormality recognition on challenging and randomly sampled cases, using 300 samples from each subset to examine the effectiveness of selective regional distillation. As shown in Table~\ref{tab:region_qa}, our method consistently improves precision, recall, and F1 across all four target abnormalities in both subsets. The gains are particularly pronounced on challenging cases, demonstrating its effectiveness in recognizing difficult abnormalities. Meanwhile, the consistent improvements on randomly sampled cases indicate that these benefits extend beyond challenging cases to more general regional recognition.

\begin{figure*}[t]
    \centering
    \includegraphics[width=\textwidth]{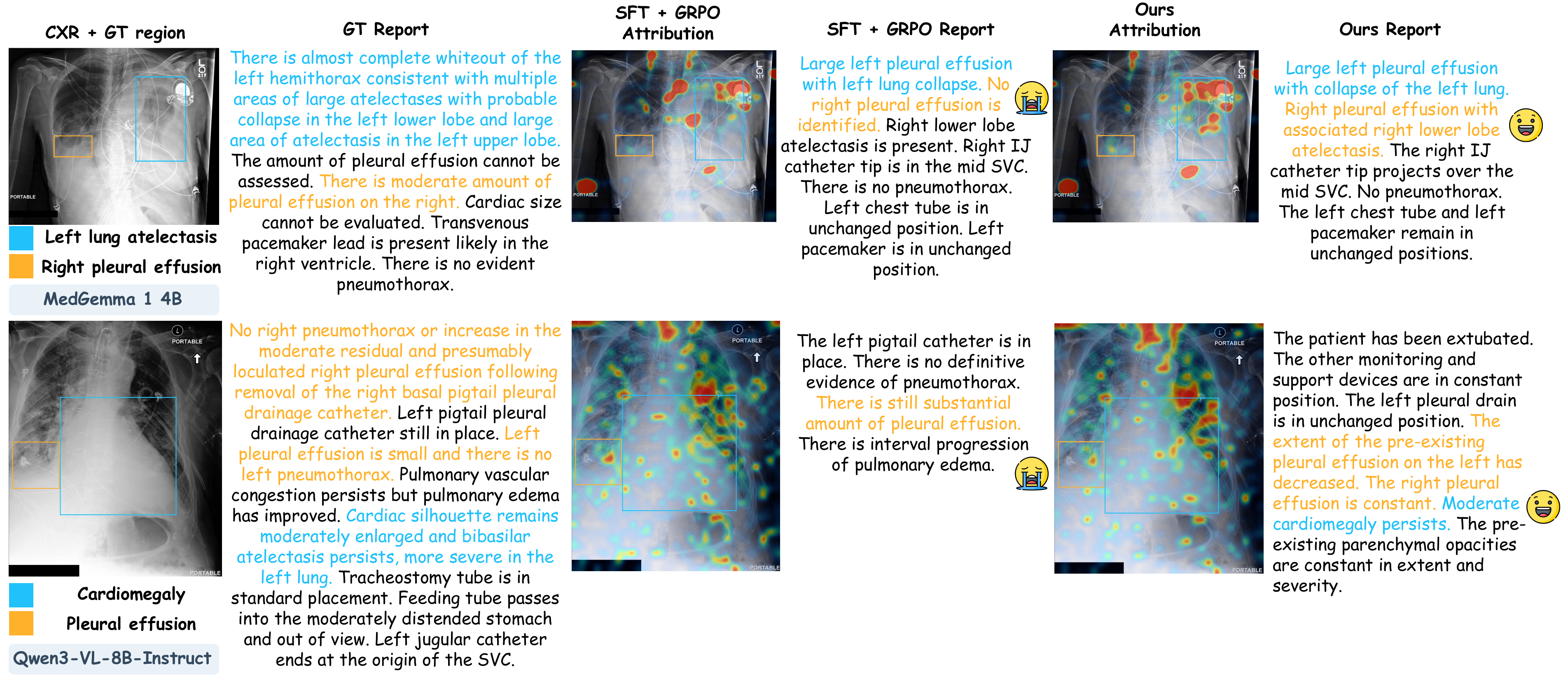}
    \caption{\textbf{Qualitative comparison of visual attribution before and after regional distillation.}
    Representative error-correction cases from MedGemma 1 4B (top) and Qwen3-VL-8B-Instruct (bottom), showing changes in generated reports and visual attribution to the corresponding abnormality regions.
    Attribution maps are computed using gradient $\times$ activation. For Qwen3-VL, attributions are aggregated from the final visual embeddings and DeepStack features, whereas for MedGemma, they are computed at an intermediate visual encoder layer.
    Due to differences in model architecture, feature resolution, and visualization settings, the attribution maps are computed differently and may exhibit distinct appearances. Therefore, we focus on before-and-after comparisons within each model rather than directly comparing attribution intensities across models.}
    \label{fig:qualitative_analysis}
    \vspace{-15pt}
\end{figure*}

\begin{table}[t]
\centering
\caption{Region-level abnormality recognition on challenging
and randomly sampled cases from MIMIC-CXR.
Each subset contains 300 samples: the challenging subset
consists of 150 positive and 150 negative cases, while
the random subset is randomly sampled.
P, R, and F1 denote precision, recall, and F1 score,
respectively.}
\label{tab:region_qa}
\setlength{\tabcolsep}{3pt}
\renewcommand{\arraystretch}{1.1}
\resizebox{\columnwidth}{!}{
\begin{tabular}{cccccccc}
\toprule
\multirow{2}{*}{Abnormality} &
\multirow{2}{*}{Method} &
\multicolumn{3}{c}{Challenging Samples} &
\multicolumn{3}{c}{Random Samples} \\
\cmidrule(lr){3-5}
\cmidrule(lr){6-8}
& & P & R & F1 & P & R & F1 \\
\midrule
\multicolumn{1}{c|}{\multirow{2}{*}{Atelectasis}}
& SFT + GRPO & 30.91 & 34.00 & 32.38 & 64.79 & 83.64 & 73.02 \\
\multicolumn{1}{c|}{}
& \cellcolor{aliceblue!60}\textbf{Ours}
& \cellcolor{aliceblue!60}52.78
& \cellcolor{aliceblue!60}63.33
& \cellcolor{aliceblue!60}57.58
& \cellcolor{aliceblue!60}70.07
& \cellcolor{aliceblue!60}87.27
& \cellcolor{aliceblue!60}77.73 \\
\midrule
\multicolumn{1}{c|}{\multirow{2}{*}{Consolidation}}
& SFT + GRPO & 27.85 & 29.33 & 28.57 & 36.67 & 73.33 & 48.89 \\
\multicolumn{1}{c|}{}
& \cellcolor{aliceblue!60}\textbf{Ours}
& \cellcolor{aliceblue!60}36.36
& \cellcolor{aliceblue!60}50.66
& \cellcolor{aliceblue!60}42.34
& \cellcolor{aliceblue!60}41.38
& \cellcolor{aliceblue!60}80.00
& \cellcolor{aliceblue!60}54.55 \\
\midrule
\multicolumn{1}{c|}{\multirow{2}{*}{Cardiomegaly}}
& SFT + GRPO & 26.21 & 25.33 & 25.76 & 62.39 & 73.17 & 67.35 \\
\multicolumn{1}{c|}{}
& \cellcolor{aliceblue!60}\textbf{Ours}
& \cellcolor{aliceblue!60}39.58
& \cellcolor{aliceblue!60}38.00
& \cellcolor{aliceblue!60}38.77
& \cellcolor{aliceblue!60}66.67
& \cellcolor{aliceblue!60}78.05
& \cellcolor{aliceblue!60}71.91 \\
\midrule
\multicolumn{1}{c|}{\multirow{2}{*}{Pleural Effusion}}
& SFT + GRPO & 22.92 & 22.00 & 22.45 & 64.58 & 86.11 & 73.81 \\
\multicolumn{1}{c|}{}
& \cellcolor{aliceblue!60}\textbf{Ours}
& \cellcolor{aliceblue!60}37.58
& \cellcolor{aliceblue!60}41.33
& \cellcolor{aliceblue!60}39.37
& \cellcolor{aliceblue!60}68.56
& \cellcolor{aliceblue!60}88.88
& \cellcolor{aliceblue!60}77.41 \\
\bottomrule
\end{tabular}
}
\end{table}

We further examine whether improved regional recognition translates into more accurate clinical findings in generated reports. As shown in Table~\ref{tab:report_findings}, our method improves precision, recall, and F1 across all four target abnormalities, with particularly notable gains in consolidation. Consolidation is relatively underrepresented in the test set and can exhibit radiographic patterns similar to other abnormalities, such as atelectasis, making its recognition challenging. The SFT- and GRPO-trained model achieves an F1 of 0 for consolidation, whereas our method increases it to 37.19. This substantial improvement suggests that selective regional distillation helps the model recognize previously missed abnormalities and incorporate them into generated reports. Meanwhile, the performance on the remaining ten findings is largely preserved, with F1 changing only slightly from 62.36 to 62.61. These results indicate that our method improves the reporting of targeted abnormalities while maintaining clinical performance on non-target findings in the generated reports.

\begin{table}[t]
\centering
\caption{Report-level clinical performance on MIMIC-CXR.
Disease labels are extracted from generated and reference
reports using CheXbert. Count denotes the number of
positive reference cases for each finding in the MIMIC-CXR
test set. Other 10 Findings denotes the remaining ten
categories not targeted by regional distillation, with
their results micro-averaged and reported.}
\label{tab:report_findings}
\setlength{\tabcolsep}{4pt}
\renewcommand{\arraystretch}{1.1}
\resizebox{\columnwidth}{!}{
\begin{tabular}{cccccc}
\toprule
Finding & Count & Method & Precision & Recall & F1 \\
\midrule
\multirow{2}{*}{Atelectasis}
& \multicolumn{1}{c|}{\multirow{2}{*}{775}}
& SFT + GRPO & 45.37 & 65.16 & 53.50 \\
& \multicolumn{1}{c|}{}
& \cellcolor{aliceblue!60}\textbf{Ours}
& \cellcolor{aliceblue!60}49.64
& \cellcolor{aliceblue!60}70.84
& \cellcolor{aliceblue!60}58.37 \\
\midrule
\multirow{2}{*}{Consolidation}
& \multicolumn{1}{c|}{\multirow{2}{*}{199}}
& SFT + GRPO & 0.00 & 0.00 & 0.00 \\
& \multicolumn{1}{c|}{}
& \cellcolor{aliceblue!60}\textbf{Ours}
& \cellcolor{aliceblue!60}44.14
& \cellcolor{aliceblue!60}32.16
& \cellcolor{aliceblue!60}37.20 \\
\midrule
\multirow{2}{*}{Cardiomegaly}
& \multicolumn{1}{c|}{\multirow{2}{*}{830}}
& SFT + GRPO & 52.39 & 62.17 & 56.86 \\
& \multicolumn{1}{c|}{}
& \cellcolor{aliceblue!60}\textbf{Ours}
& \cellcolor{aliceblue!60}56.40
& \cellcolor{aliceblue!60}68.43
& \cellcolor{aliceblue!60}61.84 \\
\midrule
\multirow{2}{*}{Pleural Effusion}
& \multicolumn{1}{c|}{\multirow{2}{*}{845}}
& SFT + GRPO & 65.18 & 77.75 & 70.91 \\
& \multicolumn{1}{c|}{}
& \cellcolor{aliceblue!60}\textbf{Ours}
& \cellcolor{aliceblue!60}69.04
& \cellcolor{aliceblue!60}82.36
& \cellcolor{aliceblue!60}75.11 \\
\midrule
\multirow{2}{*}{Other 10 Findings}
& \multicolumn{1}{c|}{\multirow{2}{*}{3495}}
& SFT + GRPO & 67.20 & 58.17 & 62.36 \\
& \multicolumn{1}{c|}{}
& \cellcolor{aliceblue!60}\textbf{Ours}
& \cellcolor{aliceblue!60}67.91
& \cellcolor{aliceblue!60}58.08
& \cellcolor{aliceblue!60}62.61 \\
\bottomrule
\end{tabular}
}
\end{table}

As shown in Fig.~\ref{fig:qualitative_analysis}, we further examine how selective regional distillation affects visual attribution and generated reports. In the MedGemma case (top), the original model shows limited attribution to the right pleural effusion region (orange box) and incorrectly reports its absence. After distillation, the model exhibits stronger attribution to this region and correctly identifies the pleural effusion. Similarly, in the Qwen3-VL case (bottom), the distilled model shows enhanced attribution to the cardiomegaly region (blue box), while correcting the omitted cardiomegaly and providing a more accurate description of the pleural effusion. These examples suggest that, unlike the largely unchanged visual attribution observed after GRPO alone, our method can improve clinical predictions alongside more pronounced attribution to the corresponding abnormality regions.

\section{Conclusion}

In this paper, we propose a selective regional distillation framework to enhance visual representations for chest X-ray radiology report generation (RRG). Motivated by the limited visual representation improvements observed after GRPO, our method selectively transfers verified regional features from a region-aware teacher through a Spatially Gated Residual Adapter, while preserving existing visual information. Extensive experiments demonstrate improved clinical report generation performance across datasets and MLLM backbones, particularly in clinical accuracy. Further regional recognition and visual attribution analyses support the effectiveness of our method in enhancing the recognition of challenging abnormalities. 
{
    \small
    \bibliographystyle{ieeenat_fullname}
    \bibliography{main}
}

% WARNING: do not forget to delete the supplementary pages from your submission 
% \input{sec/X_suppl}
% \input{sec/6_suppl}
\end{document}